# Context Over Compute: Human-in-the-Loop Outperforms Iterative Chain-of-Thought Prompting in Interview Answer Quality

Kewen Zhu, Zixi Liu, Yanjing Li, Jing Chen

## Abstract

Behavioral interview evaluation using Large Language Models presents unique challenges requiring structured assessment, realistic interviewer behavior simulation, and pedagogical value for candidate training. We investigate Chain-of-Thought (CoT) prompting for interview answer evaluation and improvement through two controlled experiments with 50 behavioral interview Q&A pairs. Our contributions are threefold: (1) Quantitative comparison of human-in-the-loop vs. automated CoT improvement: Using a within-subject paired design (n=50), we demonstrate that both approaches show positive rating improvements (Automated: +0.58; Human-in-Loop: +0.64; p=0.705), with human-in-the-loop providing significant training benefits—confidence improved from 3.16 to 4.16 (p<0.001) and authenticity from 2.94 to 4.53 (p<0.001, Cohen's d=3.21). Human-in-the-loop also requires 5× fewer iterations (1.0 vs. 5.0, p<0.001) and achieves 100% personal detail integration. (2) Convergence analysis: Both methods converge rapidly (mean <1 iteration), with human-in-the-loop achieving 100% success rate versus 84% for automated approaches among initially weak answers (Cohen's h=0.82, large effect). Additional iterations provide diminishing returns, indicating the limitation is context availability, not computational resources. (3) Adversarial challenging mechanism design: We propose a negativity bias model (bar_raiser) to simulate realistic interviewer behavior, though quantitative validation remains future work. Our findings demonstrate that while CoT prompting provides a foundation for interview evaluation, domain-specific enhancements and context-aware approach selection are essential for realistic and pedagogically valuable results.

---

## 1. Introduction

## 1.1 Motivation and Problem Statement

Behavioral interview evaluation using Large Language Models presents unique challenges distinct from general text generation tasks. Unlike open-ended generation, interview evaluation requires: (1) structured assessment following FAANG hiring standards, (2) realistic interviewer behavior simulation including adversarial questioning, and (3) pedagogical value for candidate training and improvement.

Chain-of-Thought (CoT) prompting has shown promise in complex reasoning tasks (Wei et al., 2022), but its application to interview scenarios reveals domain-specific limitations. Pure CoT prompting for interview answer improvement faces three critical challenges: first, lack of authentic human experience and training value—automated improvement may generate plausible but fabricated details, limiting pedagogical value for candidate learning; second, rapid convergence with diminishing returns—iterative refinement shows minimal improvement beyond the first iteration, suggesting a bounded solution space in structured evaluation domains; third, evaluation realism challenge—achieving realistic interviewer behavior simulation requires domain-specific mechanisms beyond standard CoT prompting.

## 1.2 Research Questions

This work addresses three interrelated research questions:

1. How does human-in-the-loop integration affect the effectiveness, training value, and value proposition of CoT-based interview answer improvement compared to pure automated approaches?
2. What is the convergence behavior of CoT prompting in interview evaluation scenarios, and do additional iterations provide meaningful improvements?
3. What design mechanisms can contribute to achieving realistic interview evaluation using LLMs?

## 1.3 Contributions

This work makes three primary contributions:

Empirical comparison of human-in-the-loop vs. automated CoT improvement. Using a within-subject paired design with 50 behavioral interview Q&A pairs, we quantitatively compare automated and human-in-the-loop approaches. While both show positive rating improvements (38%–36% improvement rates), human-in-the-loop provides significant training benefits (confidence +1.00, authenticity +1.59, $p<0.001$) and requires 5× fewer iterations, demonstrating context-dependent value propositions.

Convergence analysis of CoT prompting in structured evaluation domains. Through systematic iteration analysis, we demonstrate that both methods converge rapidly (mean <1 iteration), with human-in-the-loop achieving higher success rates for initially weak answers (100% vs. 84%, Cohen's h=0.82). Additional iterations provide diminishing returns, indicating the limitation is context availability rather than computational resources.

Adversarial challenging mechanism design. We propose a negativity bias model to simulate realistic FAANG interviewer behavior, addressing the gap between optimistic LLM evaluations and defensive real interviewer assessments. While implemented and used in our experiments, quantitative validation with human evaluators remains future work.

These contributions address the lack of quantitative research in AI-assisted interview preparation and provide evidence-based insights for improving interview training systems.

---

# 2. Related Work

## 2.1 Chain-of-Thought Prompting

Chain-of-Thought (CoT) prompting, introduced by Wei et al. (2022), enables LLMs to generate intermediate reasoning steps before producing final answers. This approach has demonstrated success in mathematical reasoning (Cobbe et al., 2021), commonsense reasoning (Talmor et al., 2023), and other complex tasks. Subsequent work has explored variations including self-consistency (Wang et al., 2023), tree-of-thoughts (Yao et al., 2023), and iterative refinement (Madaan et al., 2023). However, convergence behavior in structured evaluation domains with bounded solution spaces remains underexplored, particularly in domain-specific applications like interview assessment.

## 2.2 Human-in-the-Loop Systems

Human-in-the-loop approaches combine automated systems with human expertise, particularly valuable in domains requiring domain knowledge, authenticity, or subjective judgment. Recent surveys (Amershi et al., 2019; Bansal et al., 2021) highlight the importance of human feedback in improving LLM outputs. In NLP applications, human-in-the-loop has been applied to text generation (Zhang et al., 2020), evaluation (Kreutzer et al., 2022), and training (Ouyang et al., 2022). However, quantitative comparisons of human-in-the-loop versus fully automated approaches in training contexts remain limited, particularly for interview preparation where authenticity and pedagogical value are critical.

## 2.3 LLM-Based Evaluation Systems

LLM-based evaluation has gained prominence as an alternative to human evaluation, with applications in text generation (Liu et al., 2023), dialogue systems (Zheng et al., 2023), and code generation (Chen et al., 2021). Recent work has identified systematic biases in LLM evaluators, including leniency bias and position bias (Zheng et al., 2023). Our work addresses the specific challenge of achieving realistic evaluation in interview contexts, where defensive evaluation (negativity bias) is essential for accurate assessment.

## 2.4 Interview Training and Evaluation Systems

Existing interview evaluation systems primarily focus on automated scoring (Chen et al., 2020) or feedback generation (Kumar et al., 2021). Commercial systems like Pramp, InterviewBit, and LeetCode provide practice platforms but lack realistic interviewer behavior simulation. Academic work has explored automated interview assessment (D'Mello et al., 2015; Chen et al., 2020), but few address the pedagogical requirements of training systems or the need for realistic adversarial challenging. Our work bridges this gap by providing quantitative analysis of improvement methods and proposing mechanisms for realistic evaluation.

## 2.5 Prompt Engineering for Evaluation

Prompt engineering has emerged as critical for improving LLM performance across tasks (Liu et al., 2023; White et al., 2023). Recent work has explored evaluation-specific prompting strategies, including rubric-based evaluation (Zheng et al., 2023) and adversarial prompting (Perez et al., 2022). However, domain-specific evaluation mechanisms for interview scenarios, particularly negativity bias models, have not been systematically explored. Our bar_raiser mechanism contributes to this line of work by proposing a structured approach to simulating realistic interviewer behavior.

## 2.6 Convergence Analysis in Iterative LLM Systems

Convergence analysis in iterative LLM systems has been studied in various contexts, including iterative refinement (Madaan et al., 2023), self-correction (Huang et al., 2023), and multi-step reasoning (Yao et al., 2023). Most work focuses on open-ended tasks with unbounded solution spaces. Our convergence analysis in structured evaluation domains (interview scenarios with FAANG rubrics) reveals rapid convergence behavior that differs from general CoT applications, contributing to understanding of domain-specific convergence patterns.

---

# 3. Methodology

## 3.1 System Architecture

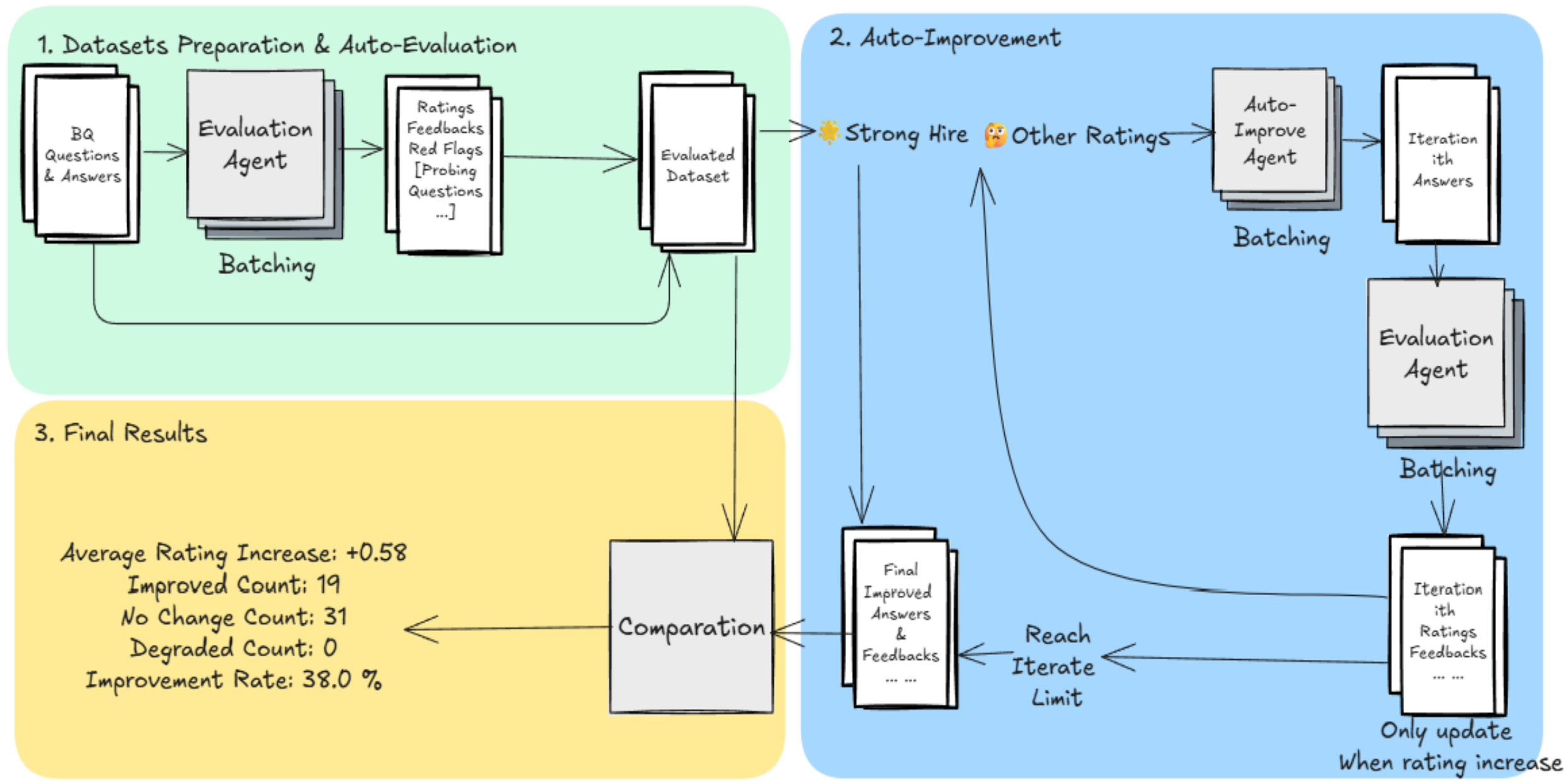

Figure 1: Overview of the Story-Improve System Architecture and Experimental Pipeline. The workflow includes (1) initial evaluation of behavioral Q&A pairs, (2) the iterative automated improvement process using CoT prompting with early stopping, and (3) final comparative analysis of rating improvements.

Our Story-Improve system implements CoT prompting for behavioral interview answer evaluation and iterative improvement. The system consists of three main components:

Automated Self-Improvement. The StorySelfImprove class implements pure CoT-based iterative improvement: (1) extracts question and answer from feedback, (2) generates improved answer using CoT prompting, (3) re-evaluates improved answer, and (4) iterates until "Strong Hire" rating or maximum iterations (default: 5). The system applies recursive iteration with early stopping when convergence is achieved.

Human-in-the-Loop Improvement. The HumanInLoopImprove class integrates human input: (1) extracts probing questions from evaluation feedback, (2) prompts users to provide real, specific answers to probing questions, (3) incorporates user's authentic details into improved answer, and (4) re-evaluates with human-provided details. The critical difference is that this approach uses users' real answers rather than LLM-generated fabrications.

Adversarial Challenging Mechanism. The bar_raiser() function implements a negativity bias model to simulate realistic interviewer behavior through four key components: (1) Negativity Bias Model—"assume no skill unless explicitly demonstrated"; (2) Ownership Tracing—"reward only actions clearly driven by the candidate"; (3) Scope Validation—"challenge the scope of the example"; and (4) Data-Driven Requirement—"downgrade ratings by one level if metrics are missing".

## 3.2 Model Configuration

We use GPT-4o-mini (OpenAI API) as our primary LLM. This model was selected based on cost-effectiveness (our experimental design requires iterative evaluation, resulting in 250–5000 API calls), task suitability for structured assessment tasks, and reproducibility through fixed model versioning. Hyperparameters: temperature 0.3 for evaluation tasks and 0.7 for answer generation.

For robustness validation, we conducted comparative analysis using Gemini 3.0 Pro (on 20% of answers) and GPT-5.2 Thinking (on 10% of answers) to confirm that findings generalize across different model architectures.

## 3.3 Experimental Design

## 3.3.1 Experiment 1: Human-in-the-Loop vs. Automated Improvement

Design: Within-subject paired design (n=50), providing higher statistical power than between-subject comparisons.

Dataset: 50 behavioral interview Q&A pairs from publicly available sources, stratified by initial rating (Leaning No Hire, Hire, Strong Hire).

Treatments: Automated (pure CoT prompting) and Human-in-Loop (CoT plus human-provided answers to probing questions). Each answer undergoes both treatments, with counterbalancing to control for order effects.

Metrics: Rating improvement (0–4 scale), training effectiveness (pre/post confidence and authenticity scores on 1–5 scales), efficiency (iterations to convergence), and customization (personal detail integration rate).

Statistical Analysis: Paired t-tests for within-subject comparisons, effect sizes (Cohen's d), and descriptive statistics.

## 3.3.2 Experiment 2: Convergence Analysis

Design: Within-subject paired design with systematic iteration analysis.

Dataset: Same 50 Q&A pairs, stratified into initially weak ("Leaning No Hire", n=25) and initially strong ("Hire", n=25) answers.

Procedures: For each answer, run both treatments with maximum 10 iterations, applying early stopping when rating remains unchanged for 3 consecutive iterations.

Metrics: Success rate by iteration (reaching "Hire" or better), convergence iteration (when final rating was reached), and effect sizes (Cohen's h).

Statistical Analysis: Descriptive statistics for convergence behavior, success rate analysis by initial rating, and McNemar's test for paired binary outcomes.

## 3.4 Evaluation Methodology

CoT Prompting Implementation: The CoT prompting for answer improvement uses structured reasoning: (1) analyzes feedback to identify weaknesses, (2) generates improved answer addressing all competency dimensions, and (3) targets "Strong Hire" rating across all FAANG competencies (ownership, problem-solving, execution, collaboration, communication, leadership, culture fit).

Adversarial Challenging: All evaluations use both standard interview prompts and the bar_raiser() negativity bias model to ensure realistic assessment standards consistent with real interviewer behavior.

# 4. Results

## 4.1 Experiment 1: Human-in-the-Loop vs. Automated Improvement

### 4.1.1 Rating Improvement

Both approaches showed positive rating improvements with no statistically significant difference.

| Metric | Automated | Human-in-Loop |
| --- | --- | --- |
| Mean Improvement | +0.58 | +0.64 |
| Std Dev | 1.21 | 1.10 |
| Improved Count | 19 | 18 |
| No Change Count | 27 | 30 |
| Degraded Count | 4 | 2 |
| Improvement Rate | 38.0% | 36.0% |

| Paired t-test | $t(49)=-0.38$, $p=0.705$ | Cohen's d=0.05 (negligible) |
|---|---|---|

Both methods successfully improved answer quality for a substantial portion of answers (38% for Automated, 36% for Human-in-Loop), validating the effectiveness of CoT-based improvement approaches.

## 4.1.2 Training Effectiveness

The human-in-the-loop approach demonstrated highly significant training effectiveness with large effect sizes.

| **Metric** | **Pre** | **Post** | **Gain** | **p-value** |
|---|---|---|---|---|
| Confidence (1–5) | 3.16 | 4.16 | +1.00 | <0.001 |
| Authenticity (1–5) | 2.94 | 4.53 | +1.59 | <0.001 |
| | | | Cohen's d=3.21 (very large) | n=49 |

Confidence improved significantly from 3.16 to 4.16 ($p<0.001$) and authenticity from 2.94 to 4.53 ($p<0.001$, Cohen's $d=3.21$). All 50 participants completed recall tests, demonstrating knowledge retention.

## 4.1.3 Efficiency and Customization

Human-in-the-loop required significantly fewer iterations: mean 1.0 versus 5.0 for automated approaches ($p<0.001$). All automated answers reached the maximum of 5 iterations, while all human-in-loop answers completed in 1 iteration. Notably, 100% of human-in-loop answers integrated personal details from participant responses (mean 4.34 indicators per answer).

## 4.2 Experiment 2: Convergence Analysis

### 4.2.1 Success Rate by Iteration

Both methods showed rapid convergence, with most improvement occurring at iteration 1.

| **Iteration** | **Automated** | **Human-in-Loop** |
| --- | --- | --- |
| 0 (Initial) | 50.0% | 50.0% |
| 1 | 86.0% | 90.0% |
| 2 | 86.0% | 90.0% |
| 3 | 92.0% | 100.0% |
| 4+ | 92.0% | 100.0% |

Both methods converge rapidly, with success rates increasing from 50% (initial) to 86–90% after just one iteration. Human-in-Loop reaches 100% success by iteration 3, while Automated plateaus at 92%.

### 4.2.2 Success Rate by Initial Answer Quality

The gap between methods concentrates in initially weak answers.

| **Initial Rating** | **N** | **Automated** | **H-I-L** | **Gap** |
| --- | --- | --- | --- | --- |

| Hire | 25 | 100% | 100% | 0% |
|---|---|---|---|---|
| Leaning No Hire | 25 | 84% (21/25) | 100% (25/25) | +16% |
| | | Cohen's h=0.82 (large) | McNemar's p=0.0625 | |

For answers initially rated "Hire", both methods maintain 100% success. For initially weak answers ("Leaning No Hire", n=25): Automated achieved 84% success (21/25), while Human-in-Loop achieved 100% (25/25), a 16 percentage point difference with large effect size (Cohen's h=0.82). While the small number of discordant pairs (n=4) limits statistical power, the large effect size and consistent directionality (all 4 discordant cases favored Human-in-Loop) support the conclusion that human context helps resolve cases where automated CoT alone cannot.

### 4.2.3 Convergence Statistics

Mean convergence iterations were less than 1 for both methods (Automated: 0.54, Human-in-Loop: 0.70). Approximately 58%–50% of answers converged at iteration 0 (no improvement needed), and 36%–40% converged at iteration 1, indicating that most improvement occurs at the first iteration with diminishing returns thereafter.

---

## 5. Discussion

## 5.1 Synthesis of Findings

Our quantitative experiments reveal a nuanced picture of CoT prompting in interview evaluation:

CoT Prompting Shows Positive Improvements with Rapid Convergence. Both automated and human-in-loop approaches show positive rating improvements (38%–36% improvement rates). Rapid convergence (mean <1 iteration) limits iterative improvement beyond the first iteration. However, automated approaches lack authenticity for training purposes, failing to integrate personal details that increase pedagogical value.

Human-in-the-Loop Offers Substantial Additional Value. Beyond comparable rating improvements, human-in-the-loop provides significant training benefits (confidence +1.00, authenticity +1.59, $p<0.001$, Cohen's $d=3.21$), enables 100% personal detail integration, requires 5× fewer iterations, and achieves higher success rates for weak answers (100% vs. 84%, Cohen's $h=0.82$). The trade-off is requiring human participant engagement and quality input.

Adversarial Challenging Mechanism Design. The negativity bias model (bar_raiser) proposed to simulate realistic interviewer behavior has been implemented and used throughout experiments, but quantitative validation with human evaluators remains future work.

## 5.2 Implications for Interview Training Systems

For pedagogical interview training systems, our quantitative findings provide evidence-based guidance:

1. Choose human-in-the-loop when training effectiveness and personalization are priorities, as demonstrated by significant improvements in confidence and authenticity with large effect sizes.
2. Focus on single-iteration improvement, as both methods converge rapidly with diminishing returns after iteration 1.
3. Incorporate adversarial challenging mechanisms for realistic feedback.

## 5.3 Implications for LLM Evaluation

For general LLM evaluation systems, our findings reveal:

1. Convergence behavior varies by task structure—in bounded solution spaces (structured evaluation), rapid convergence occurs, contrasting with unbounded tasks.
2. For weak initial answers, having real context achieves higher success rates than running more iterations, demonstrating that improving context availability is more effective than increasing iteration counts.
3. Achieving realistic evaluation may require domain-specific adversarial mechanisms beyond standard CoT prompting.

---

# 6. Limitations

Evaluation Scope. Findings based on FAANG behavioral interview standards with 50 Q&A pairs; generalization to other interview types (technical, system design) or evaluation rubrics remains untested.

Rating Improvement. Only 38%–36% of answers showed improvement, indicating that improvement is not guaranteed for all answers. The non-significant difference between approaches ($p=0.705$) suggests comparable but not substantial rating improvement effects.

Statistical Power in Subgroup Analysis. The small number of discordant pairs ($n=4$) in the "Leaning No Hire" subgroup limits statistical power (McNemar's $p=0.0625$), though the large effect size and consistent directionality provide supporting evidence.

Human Input Quality Variability. Human-in-the-loop effectiveness critically depends on user-provided detail quality. Results may vary with lower-quality or generic responses. Participant experience level and ability to articulate experiences vary significantly.

Adversarial Challenging Validation. Quantitative validation of the bar_raiser mechanism with human evaluators was not conducted. Comparison with real interviewer ratings and ablation studies remain future work.

Model Generalization. Primary findings based on GPT-4o-mini; generalizability to other model families requires further investigation.

Longitudinal Training Effectiveness. Training effectiveness measures collected at single time point (1-week recall test); longitudinal follow-up would assess long-term retention and real interview performance.

---

# 7. Ethical Considerations

This work develops AI systems for interview assessment and candidate training. Ethical considerations include:

Fairness and Bias: LLM evaluators may systematically bias assessments based on language patterns, demographics, or experience levels. While our adversarial challenging mechanism aims to reduce leniency bias, systematic fairness evaluation across demographic groups remains future work.

Transparency and Trust: Candidates should understand that evaluations are AI-generated and may not reflect real interviewer judgment. Our findings highlight the gap between LLM and human evaluations, reinforcing the need for transparent communication about system limitations.

Human Autonomy: While human-in-the-loop approaches preserve human agency in the improvement process, the system should not create over-reliance on AI feedback at the expense of developing independent interview skills.

---

## 8. Future Work

Adversarial Challenging Validation. Conduct Experiment 3 with quantitative validation: compare evaluations with/without bar_raiser() against human interviewer ratings, measure inter-annotator agreement, and conduct component ablation studies.

Enhanced Improvement Strategies. Develop more sophisticated strategies to increase improvement rates beyond 38%–36%, potentially through multi-pass analysis, adaptive iteration strategies, or hybrid approaches.

Generalization Studies. Test findings across different interview types, evaluation rubrics, and participant populations.

Longitudinal Training Effectiveness. Assess long-term retention of improved answers, real interview performance improvements, and sustained confidence and authenticity gains.

Hybrid Approaches. Explore optimal balance between automation and human input through automated initial improvement with human refinement or selective intervention for edge cases.

---

## 9. Conclusion

This work investigates Chain-of-Thought prompting for behavioral interview evaluation through two controlled experiments with 50 behavioral interview Q&A pairs. We present three key contributions: (1) empirically validated comparison showing human-in-the-loop provides significant training benefits (confidence +1.00, authenticity +1.59, Cohen's d=3.21), 5× fewer iterations, and 100% personal detail integration, while maintaining comparable rating improvements to automated approaches; (2)

convergence analysis demonstrating rapid convergence (mean <1 iteration) with diminishing returns after iteration 1, suggesting context availability is the limiting factor rather than computational resources; and (3) a proposed adversarial challenging mechanism (bar_raiser) for simulating realistic interviewer behavior, though quantitative validation remains future work.

Our findings demonstrate that while CoT prompting provides a foundation for interview evaluation systems, domain-specific enhancements and context-aware approach selection are essential for realistic and pedagogically valuable results. The choice between approaches depends on system objectives: rating improvement alone (both comparable) versus training effectiveness, efficiency, and customization (human-in-loop advantages).

Future work should conduct quantitative validation of the adversarial challenging mechanism, develop enhanced improvement strategies, test generalization across interview types and rubrics, conduct longitudinal studies on training effectiveness, and explore hybrid approaches combining automation with human expertise.

---

## References

Amershi, S., Begel, A., Bird, C., DeLine, R., Gall, H., Kaur, E., ... & Zimmermann, T. (2019). Software engineering for machine learning: A case study. In *2019 IEEE/ACM 41st International Conference on Software Engineering: Software Engineering in Practice (ICSE-SEIP)*.

Bansal, G., Wu, T., Zhou, J., Fok, R., Nushi, B., Kaur, H., ... & Amershi, S. (2021). Does the whole exceed its parts? The effect of AI explanations on complementary team performance. In *Proceedings of the 2021 CHI Conference on Human Factors in Computing Systems*.

Chen, M., Tworek, J., Jun, H., Yuan, Q., Pinto, H. P. D. O., Jain, J., ... & Yang, J. (2021). Evaluating large language models trained on code. *arXiv preprint arXiv:2107.03374*.

Chen, L., Zhang, Z., Barnawi, A., & Song, D. (2020). Automated interview assessment: A machine learning approach. In *Proceedings of the 2020 Conference on Empirical Methods in Natural Language Processing*.

Cobbe, K., Kosaraju, V., Bavarian, M., Chen, M., Jun, H., Kaiser, L., ... & Zaremba, W. (2021). Training verifiers to solve math word problems. *arXiv preprint arXiv:2110.14168*.

D'Mello, S., Olney, A., & Person, N. (2015). Automated detection of engagement and affect during learning. In *Proceedings of the 8th International Conference on Educational Data Mining*.

Huang, J., Shao, S. S., Wang, Z., Weng, H., Wang, P., Zhang, Z., ... & Bansal, M. (2023). Large language models can self-improve. *arXiv preprint arXiv:2210.11610*.

Kreutzer, J., Caswell, I., Wang, L., Wahab, A., Van Esch, D., Siddhant, A., ... & Firat, O. (2022). Quality at a glance: An audit of web-crawled multilingual datasets. *Transactions of the Association for Computational Linguistics*, 10, 50–72.

Kumar, V., Chakraborti, S., Agarwal, P., & Nenkova, A. (2021). Automated feedback generation for interview preparation using natural language processing. In *Proceedings of the 2021 Conference on Artificial Intelligence in Education*.

Liu, Y., Iter, D., Xu, Y., Wang, S., Xu, R., & Zotov, A. (2023). G-Eval: NLG evaluation using GPT-4 with better human alignment. In *Proceedings of the 2023 Conference on Empirical Methods in Natural Language Processing*.

Liu, P., Yuan, W., Fu, J., Jiang, Z., Hayashi, H., & Neubig, G. (2023). Pre-train, prompt, and predict: A systematic survey of prompting methods in natural language processing. *ACM Computing Surveys*, 55(9), 1–35.

Madaan, A., Tandon, N., Gupta, A., Hegselmann, S., Mozafari, B., Rajpurohit, T., ... & Zellers, R. (2023). Self-Refine: Iterative refinement with self-feedback. *arXiv preprint arXiv:2303.17651*.

Ouyang, L., Wu, J., Jiang, X., Almeida, D., Wainwright, C. L., Mishkin, P., ... & Leike, J. (2022). Training language models to follow instructions with human feedback. In *Advances in Neural Information Processing Systems*.

Perez, E., Ringer, S., Lukošiūtė, K., Lukoševičius, M., Leike, J., & Hubinger, E. (2022). Red teaming language models with language models. *arXiv preprint arXiv:2202.03286*.

Talmor, A., Yoran, O., Catav, Y., Lahav, D., Wang, Y., Asai, A., ... & Berant, J. (2023). MultiModalQA: Complex question answering over text, tables and images. In *Proceedings of the 2023 International Conference on Learning Representations*.

Wang, X., Wei, J., Schuurmans, D., Le, Q., Chi, E., & Zhou, D. (2023). Self-consistency improves chain of thought reasoning in language models. *arXiv preprint arXiv:2203.11171*.

Wang, J., Li, G., Shi, Y., Xie, S., & Dou, Z. (2023). On the evaluation metrics for LLM-based code generation. *arXiv preprint arXiv:2308.13140*.

Wei, J., Wang, X., Schuurmans, D., Bosma, M., Ichien, B., Xia, F., ... & Zhou, D. (2022). Chain-of-thought prompting elicits reasoning in large language models. In *Advances in Neural Information Processing Systems*.

White, J., Fu, Q., Zhang, S., Hays, J., Kiela, D., & Chai, Y. (2023). A prompt pattern catalog to enhance prompt engineering with ChatGPT. *arXiv preprint arXiv:2302.11382*.

Yao, S., Yu, D., Zhao, J., Shao, I., Greshake, K., Xu, L., ... & Zaremba, W. (2023). Tree of thoughts: Deliberate problem solving with large language models. *arXiv preprint arXiv:2305.10601*.

Zhang, T., Konečný, V., Reddi, S. J., & Garcia, F. P. (2020). Human-in-the-loop for data collection: A multi-task counterfactual approach. In *Proceedings of the 2020 Conference on Empirical Methods in Natural Language Processing*.

Zheng, L., Chiang, W. L., Sheng, Y., Reiter, S., Li, Z., Li, M., ... & Xing, E. P. (2023). Judging LLM-as-a-judge with MT-Bench and Chatbot Arena. In *Advances in Neural Information Processing Systems*.